\documentclass[journal]{IEEEtran}

\IEEEoverridecommandlockouts
\usepackage{cite}
\usepackage{amsmath,amssymb,amsfonts}
\usepackage{algorithmic}
\usepackage{graphicx}
\usepackage{textcomp}
\usepackage{xcolor}
\usepackage{caption}

\usepackage{tabularx}
\usepackage{booktabs}
\usepackage{multirow}
\usepackage{float}       
\usepackage{placeins}    
\usepackage{dblfloatfix}

\def\BibTeX{{\rm B\kern-.05em{\sc i\kern-.025em b}\kern-.08em
    T\kern-.1667em\lower.7ex\hbox{E}\kern-.125emX}}
\begin{document}

\title{DeepTokenEEG: Enhancing Mild Cognitive Impairment and Alzheimer’s Classification via Tokenized EEG Features}

\author{
    \IEEEauthorblockN{Thinh Nguyen-Quang\textsuperscript{1}},
    \IEEEauthorblockN{Minh Long Ngo\textsuperscript{1}},
    \IEEEauthorblockN{Ngoc-Son Nguyen\textsuperscript{1}},
    \IEEEauthorblockN{Nguyen Thanh Vinh\textsuperscript{2}},
    \IEEEauthorblockN{Huy-Dung Han\textsuperscript{1}}, \\
    \IEEEauthorblockN{Bui Thanh Tung\textsuperscript{2}},
    \IEEEauthorblockN{Nguyen Quang Linh\textsuperscript{3}},
    \IEEEauthorblockN{Khuong Vo\textsuperscript{4}},
    \IEEEauthorblockN{Manoj Vishwanath\textsuperscript{4,*}},
    \IEEEauthorblockN{Hung Cao\textsuperscript{4,*}}

    \thanks{\textsuperscript{1}Thinh Nguyen-Quang, Minh Long Ngo, Ngoc Son Nguyen and Huy-Dung Han are with Hanoi University of Science and Technology, Hanoi, Vietnam (e-mail: nqthinh.airesearch@gmail.com; long.nm223717@sis.hust.edu.vn; son.nn213771@sis.hust.edu.vn; dung.hanhuy@hust.edu.vn).}
    \thanks{\textsuperscript{2}Nguyen Thanh Vinh and Bui Thanh Tung are with VNU University of Engineering and Technology, Hanoi, Vietnam (e-mail: \{21020710, hanv, tungbt\}@vnu.edu.vn).}
    \thanks{\textsuperscript{3}Nguyen Quang Linh is with Central Military Hospital 108, Hanoi, Vietnam (e-mail: dr.linhnguyenquang@gmail.com).}
    \thanks{\textsuperscript{4}Khuong Vo, Manoj Vishwanath and Hung Cao are with University of California, Irvine, California, USA (e-mail:\{khuongav, manojv, hungcao\}@uci.edu).}
    \thanks{\textsuperscript{*}These authors contributed equally to this work and are co-corresponding authors.}
}

\maketitle

\begin{abstract}
    The detection of Alzheimer’s disease (AD) is considered crucial, as timely intervention can improve patient outcomes. Electroencephalogram (EEG)-based diagnosis has been recognized as a non-invasive, accessible, and cost-effective approach for AD detection; however, it faces challenges related to data availability, accuracy of modern deep learning methods, and the time-consuming nature of expert-based interpretation. In this study, a novel lightweight and high-performance model, DeepTokenEEG, was designed for the diagnosis of AD and the classification of EEG signals from AD patients, individuals with other neurological conditions, and healthy subjects. Unlike traditional heavy-weight models, DeepTokenEEG ultilizes spatial and temporal tokenizer that effectively captures AD-related biomarkers in both temporal and frequency domain with only 0.29 million paramaters. Trained in a combined dataset of 274 subjects, including 180 AD cases, and 94 healthy controls, the proposed method achieves a maximum recorded accuracy of 100\% on specific frequency bands, representing an improvement of 1.41-15.35\% over state-of-the-art methods on the same dataset. These results indicate the potential of DeepTokenEEG for early detection and screening of AD, with promising applicability for deployment due to its compact size.

\end{abstract}

\begin{IEEEkeywords}
   EEG; Tokenization; Alzheimer; Alzheimer’s disease detection; Mild cognitive impairment; Lightweight neural networks.
\end{IEEEkeywords}

\section{Introduction}
\label{introduction}
Alzheimer’s disease (AD) is a neurodegenerative disorder characterized by a progressive decline in cognitive functions, including memory, reasoning, and behavior. It affects millions of people worldwide and remains one of the leading causes of dementia\cite{PMID:1599598}, with no definitive cure currently available. Early detection of AD is crucial for timely interventions and improving patient outcomes, as it allows the initiation of treatments that can slow disease progression and improve quality of life. However, traditional diagnostic methods, such as clinical evaluations and neuroimaging techniques, often rely on subjective interpretation and are time-consuming, making early diagnosis challenging. Clinical evaluations typically include cognitive assessments and neurological examinations to assess patient reflexes, coordination, and sensory function. Neuroimaging techniques, including magnetic resonance imaging (MRI) and positron emission tomography (PET), are applied to visualize structural and functional changes in the brain caused by AD, allowing researchers to detect brain shrinkage and various types of neural activity. However, these methods are limited by subjective diagnosis of the examiner, dependence on specialized expertise, and restricted accessibility.

Electroencephalogram (EEG) signals, which reflect brain electrical activity, have shown great potential as a non-invasive tool for the detection of various neurological conditions, both in humans \cite{vishwanath2021investigation, vishwanath2025reducing, vo2022composing} and other species \cite{benomar2024wireless, vishwanath2020classification, xia2022microelectrode}. EEG is widely used in clinical settings due to its affordability, ease of use, and ability to detect abnormal brainwave patterns. Recent advancements in the field of neuroinformatics and artificial intelligence (AI) have paved the way for more efficient, automated diagnostic approaches. Inspired by breakthroughs in natural language processing (NLP) \cite{10.1145/3708036.3708089  }, where data are embedded into tokens to preserve local structure while allowing modeling entire context, we proposed a new tokenization-based method for EEG. EEG signals are transformed into EEG tokens, enabling the model to learn temporal order and long-range dependencies of the signals. By learning token representations, our method reduces sensitivity to neural artifacts and noise, enhances generalization across multiple subjects, enabling efficient modeling of long-range-dependencies. The key contributions of this approach are outlined as follows:
\begin{itemize}
    \item A novel deep-learning framework, \textbf{DeepTokenEEG}, is proposed for the identification and classification of AD from other related neurological conditions, including but not limited to mild cognitive impairment (MCI), frontotemporal dementia (FTD), and healthy controls (HC).
    \item A large-scale dataset was constructed for both AD/non-AD and multi-class classification by aggregating five publicly available datasets, comprising 373 subjects, including patients with Alzheimer’s disease (AD), mild cognitive impairment (MCI), frontotemporal dementia (FTD), Parkinson’s disease (PD), multiple sclerosis (MS), other neurological disorders, and healthy controls (HC).
    \item A lightweight DeepTokenEEG model is developed, offering high performance with strong potential for edge deployment.
    \item Comprehensive benchmarking is provided, comparing the proposed approach with state-of-the-art methods across multiple datasets and AD classification scenarios.
\end{itemize}

\section{Related Work}
\label{related_work}
\begin{table*}[ht!]
\centering
\caption{Summary of representative studies on EEG-based Alzheimer's disease (AD) and mild cognitive impairment (MCI) diagnosis}
\label{tab:related_works_summary}
\resizebox{\textwidth}{!}{
\begin{tabular}{lllll}
\toprule
\textbf{Study} & \textbf{Year} & \textbf{Method / Model} & \textbf{Features or Input} & \textbf{Main Findings / Accuracy} \\
\midrule
\multicolumn{5}{l}{\textit{(i) EEG Signal Processing and Feature Extraction}} \\[2pt]
Zhao \textit{et al.} \cite{Zhao2015-jh} & 2015 & Review study & Feature extraction and preprocessing overview & Summarized EEG preprocessing and feature methods \\
Sun and Mou \cite{Sun2023-uh} & 2023 & Review & Emerging EEG signal processing trends & Highlighted modern preprocessing and feature extraction techniques \\
Cios \textit{et al.} \cite{Aparna_UR2016-kk} & 2016 & Statistical analysis & Feature selection and extraction strategies & Demonstrated importance of feature engineering \\
Nogales and Benalc'azar \cite{Nogales2023-eh} & 2023 & Feature engineering framework & Various EEG feature types & Showed feature selection boosts performance \\
Bairagi \textit{et al.} \cite{Bairagi2018-dy} & 2018 & Supervised classifiers & Spectral and wavelet features & 94\% accuracy (AD vs. HC) \\
Youssef \textit{et al.} \cite{Youssef2021-yp} & 2021 & Network analysis + ML & Functional brain network biomarkers & 87.2\% accuracy (MCI vs. HC) \\
Siuly \textit{et al.} \cite{Siuly2020-lh} & 2020 & AR + entropy models & Permutation entropy, AR coefficients & Effective for MCI detection \\
\midrule
\multicolumn{5}{l}{\textit{(ii) Traditional Machine Learning Models}} \\[2pt]
McBride \textit{et al.} \cite{McBride2014-ng} & 2014 & SVM classifier & Statistical and spectral EEG features & 85.4\% accuracy (3-class: HC, MCI, AD) \\
Fiscon \textit{et al.} \cite{Fiscon2018-ne} & 2018 & Decision tree & Fourier and wavelet features & 91.7\% accuracy (HC vs. MCI) \\
Sharma \textit{et al.} \cite{Sharma2019-mo} & 2019 & SVM & Statistical + spectral features & 89\% accuracy (AD vs. HC) \\
Baidya \textit{et al.} \cite{Baidya2023-cp} & 2023 & Comparative study & Various traditional ML algorithms & Compared strengths/weaknesses of ML methods \\
\midrule
\multicolumn{5}{l}{\textit{(iii) Deep Learning-based Models}} \\[2pt]
Ieracitano \textit{et al.} \cite{Ieracitano2019-za} & 2019 & CNN & Power spectral density maps & 83.3\% accuracy (AD vs. HC) \\
Ieracitano \textit{et al.} \cite{Ieracitano2020-yu} & 2020 & MLP + wavelet + bispectrum & Handcrafted + spectral features & 89.2\% accuracy (AD vs. HC) \\
Bi and Wang \cite{Bi2019-zz} & 2019 & Deep Belief Network (DBN) & EEG spectrogram images & 95\% accuracy \\
Georgis-Yap \textit{et al.} \cite{Georgis-Yap2024-xi} & 2024 & Supervised/unsupervised deep models & EEG temporal patterns & Potential for AD diagnosis \\
Qureshi \textit{et al.} \cite{Qureshi2022-jg} & 2022 & Deep learning for seizure/AD & Raw EEG + embeddings & Demonstrated model generalization \\
Tang \textit{et al.} \cite{Tang2024-um} & 2024 & Transformer-based multimodal model & EEG + other modalities & Enhanced interpretability and multimodal fusion \\
\bottomrule
\end{tabular}
}
\end{table*}

AD diagnosis using EEG signals combined with ML has received increasing attention in recent years, especially in the context of early detection and low-cost solutions for low-resource settings. In this section, we classify the related literature into three main groups: (i) EEG signal processing and feature extraction techniques; (ii) traditional machine learning models; and (iii) deep learning-based models in the context of AD diagnosis. Each group is discussed in detail as follows.

\textit{i. EEG Signal Processing and Feature Extraction:} Numerous studies have focused on improving EEG signal processing techniques to enhance feature quality for diagnosing AD or mild cognitive impairment (MCI). Zhao and L.He \cite{Zhao2015-jh} provided an overview of preprocessing and feature extraction techniques, while Sun and Mou \cite{Sun2023-uh} discussed emerging trends in this domain. Commonly extracted features include entropy, spectral energy, wavelet-based descriptors, and autoregressive (AR) models. Cios et al. \cite{Aparna_UR2016-kk} and Nogales and Benalc'azar \cite{Nogales2023-eh} emphasized the pivotal role of feature selection and extraction in improving diagnostic performance. Bairagi et al. \cite{Bairagi2018-dy} used spectral and wavelet features, achieving 94\% accuracy with supervised classifiers, whereas Youssef et al. \cite{Youssef2021-yp} utilized functional brain network biomarkers to classify MCI with 87.2\% accuracy. Siuly et al. \cite{Siuly2020-lh} applied permutation entropy and AR models for MCI detection. Despite these promising results, traditional signal processing and feature engineering methods often rely on handcrafted features, which can be sensitive to noise and recording conditions. Moreover, these approaches typically require domain expertise and may not generalize well across datasets collected under different protocols.

\textit{ii. Traditional Machine Learning Models:} Machine learning approaches such as Support Vector Machines (SVM), decision trees, and k-Nearest Neighbors (k-NN) have been widely employed to classify AD and MCI based on EEG features. McBride et al. \cite{McBride2014-ng} used SVM to classify three groups (healthy, MCI, AD), achieving an accuracy of 85.4\%. Fiscon et al. \cite{Fiscon2018-ne} utilized decision trees with Fourier and wavelet features, reporting an accuracy of up to 91.7\% (HC vs. MCI). Sharma et al. \cite{Sharma2019-mo} combined multiple statistical and spectral features, using SVM to achieve nearly 89\% accuracy. In a comparative analysis, Baidya et al. \cite{Baidya2023-cp} evaluated the performance of various traditional methods, highlighting the strengths and weaknesses of each. However, traditional ML models often suffer from limited capacity in capturing complex spatiotemporal dynamics present in EEG data. Their performance is tightly coupled with the quality and relevance of extracted features, making them suboptimal when dealing with noisy or high-dimensional EEG signals.

\textit{iii. Deep Learning Models:} Ieracitano et al. \cite{Ieracitano2019-za} used power spectral density maps as input to CNNs for AD classification, achieving 83.3\% accuracy. In a follow-up study \cite{Ieracitano2020-yu}, they combined wavelet and bispectrum features with a multilayer perceptron (MLP), reaching 89.2\% accuracy. Bi and Wang \cite{Bi2019-zz} employed a deep belief network (DBN) on EEG spectrogram images, reporting an accuracy of up to 95\%. However, many of these approaches rely on signal epoching techniques, which may affect feature representation reliability. Georgis-Yap et al. \cite{Georgis-Yap2024-xi} and Qureshi et al. \cite{Qureshi2022-jg} proposed both supervised and unsupervised deep learning models for EEG analysis, initially for seizure prediction but demonstrating potential for AD diagnosis. Recently, Tang et al. \cite{Tang2024-um} introduced an enhanced Transformer-based model for multimodal AD diagnosis, enabling integration of various data modalities. Despite their superior performance, deep learning models require large labeled datasets and high computational resources for training. Additionally, their black-box nature limits interpretability, which is a critical factor in clinical decision-making. Generalizability across populations and recording conditions also remains a challenge for deploying these models in real-world settings.

Recently, a new concept has emerged that reframes EEG representation learning using tokenization—a concept originally developed in natural language processing (NLP) and later applied to other domains. Tokenization separates continuous data into discrete units, enabling models to process complex temporal or spatial patterns in sequences. In NLP, tokenization techniques such as Byte-Pair Encoding (BPE), WordPiece, and SentencePiece have improved both computational efficiency and semantic generalization \cite{song-etal-2021-fast}\cite{sennrich-etal-2016-neural}. Specifically, these techniques enhance performance by partitioning continuous data into discrete, informative units, which helps reduce signal noise and sensitivity to non-essential variations. This discretization enables the model to effectively capture long-range dependencies and focus on discriminative biomarkers while suppressing redundant background activity, thereby improving both computational efficiency and the ability to generalize across diverse subjects. Similar principles have been applied in bioinformatics\cite{brandes2022proteinbert}, where amino acid sequences are represented as biological tokens. In computer vision, image patches are converted into visual tokens as in the Vision Transformer (ViT)\cite{dosovitskiy2021vit}, and in audio modeling, spectrogram features are separated into codebook tokens for speech transformers\cite{baevski2020wav2vec}. These advances suggest that a tokenization-based framework can unify heterogeneous data representations under a shared sequence learning paradigm. In this context, EEG, both temporally and spatially structured, can benefit from token-based modeling to extract richer neural representations and enhance the interpretability of cognitive state classification.

\section{Methodology}
\label{methodology}

\begin{figure*}[h] 
    \centering
    \includegraphics[scale=0.2]{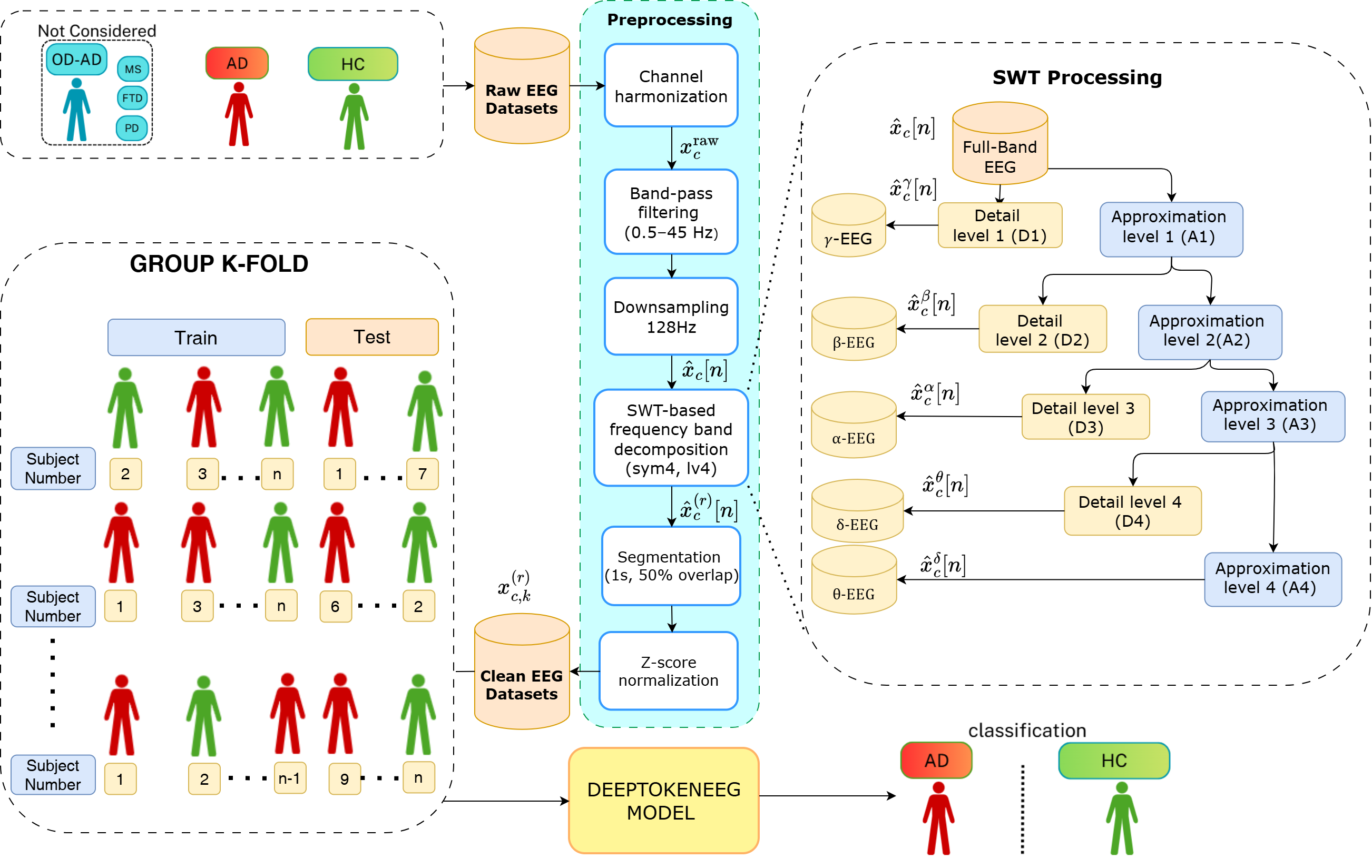}
    \caption{\small\label{fig:plot_f1_all_testset}
    The figure presents the proposed EEG-based pipeline for Alzheimer’s disease (AD) classification. Raw EEG signals from AD patients and healthy controls (HC) are first preprocessed through channel harmonization, filtering, downsampling, segmentation, and Z-score normalization. The signals are then decomposed into individual frequency bands ($\delta$, $\theta$, $\alpha$, $\beta$, and $\gamma$) using stationary wavelet transform (SWT), with each band evaluated independently to assess its discriminative power. A subject-wise cross-validation strategy is applied to prevent data leakage. Finally, the band-specific EEG representations are input to the DeepTokenEEG model for AD versus HC classification.
    }
\end{figure*}







\begin{table}[ht]
\centering
\caption{Dataset statistics with label distribution and original acquisition parameters. HC denotes cognitively healthy controls, AD Alzheimer’s disease, ODAD other AD-related diagnoses, and Unlabeled subjects without diagnostic labels.}
\label{tab:dataset_stats_with_labels}
\resizebox{0.48\textwidth}{!}{%
\begin{tabular}{lccccccccc}
\toprule
\textbf{Dataset} & \textbf{Subjects} & \textbf{HC} & \textbf{AD} &  \textbf{Channels} & \textbf{Sampling rate (Hz)}  \\
\midrule
AD-Auditory \cite{ds005048:1.0.1} 
& 35 & 10 & 17 & 19 & 250 \\

ADFSU \cite{Vicchietti_Ramos_Betting_Campanharo_2025} 
& 92 & 12 & 80 & 19 & 128\\

ADFTD \cite{ds004504:1.0.8} 
& 88 & 29 & 36 & 19 & 500\\

APAVA \cite{Smith_2017} 
& 23 & 11 & 12 & 16 & 256\\

BrainLat \cite{prado2023brainlat} 
& 135 & 32 & 35 & 128 & 512  \\

\bottomrule
\end{tabular}%
}
\end{table}
\subsection{Data aggregation} 
\label{Data aggregation}

This study focuses on EEG-based classification between individuals with AD and cognitively healthy controls (HC). Although the available datasets also include participants diagnosed with other AD-related disorders (ODAD), such as mild cognitive impairment (MCI) and frontotemporal dementia (FTD), these subjects are excluded from the present analysis, which is restricted to a binary classification between AD and HC. All datasets used in this work were acquired under resting conditions. Most datasets were recorded with participants’ eyes closed; the ADFSU dataset includes both eyes-open and eyes-closed recordings, from which only the eyes-closed data are used. In contrast, the AD-Auditory dataset is the only dataset recorded exclusively with participants’ eyes open. The detailed characteristics of each dataset, including cohort composition and recording conditions, are summarized in Table~2.

\textbf{The ADFTD dataset }\cite{ds004504:1.0.8} comprises EEG recordings from 88 participants, including 36 AD patients, 23 FTD patients, and 29 HC subjects. Data were collected using 19 scalp electrodes arranged according to the international 10–20 system, with a sampling rate of 500 Hz.

\textbf{The BrainLat dataset}\cite{prado2023brainlat} (Latin American Brain Health Institute) is a multimodal neuroimaging cohort including EEG and MRI data from five diagnostic groups: AD, behavioral-variant frontotemporal dementia (bvFTD), multiple sclerosis (MS), Parkinson’s disease (PD), and HC. In this study, we consider a subset of 135 participants with EEG data acquired using a 128-channel BioSemi ActiveTwo system, of which only the AD and HC subjects are retained for analysis.

\textbf{The AD-Auditory dataset }\cite{ds005048:1.0.1} contains EEG recordings from 35 participants, including 17 AD patients, 6 MCI cases, 10 HC subjects, and 2 unspecified cases. EEG data were acquired using 19 electrodes following the 10–20 system, with a sampling frequency of 250 Hz, and include alternating periods of baseline activity and auditory stimulation.

\textbf{The ADFSU dataset} \cite{Vicchietti_Ramos_Betting_Campanharo_2025} consists of EEG recordings from 92 participants, including 80 AD patients and 12 HC subjects. Data were recorded using 19 electrodes according to the 10–20 system, with a sampling rate of 128 Hz, and organized into short EEG segments with a trial duration of 8 seconds.

\textbf{The APAVA dataset }\cite{Smith_2017} includes EEG recordings from 23 elderly participants, comprising 12 AD patients and 11 HC subjects. Data were acquired using 16 scalp electrodes, with a sampling frequency of 256 Hz, and artifact-free EEG segments of fixed duration were pre-selected by domain experts.

In this work, we first evaluate the proposed method separately on the ADFTD and BrainLat datasets, which provide a relatively balanced distribution of AD and HC subjects. We then construct a combined multi-dataset cohort by pooling only AD and HC participants, resulting in a total of 180 AD patients and 94 healthy controls. Although this combined cohort is not perfectly balanced, it reflects the natural availability of AD cases across datasets and enables the model to learn from a diverse range of AD-related EEG patterns while maintaining a sufficient number of HC subjects for reliable evaluation.

\subsection{Data Preprocessing}
\label{Data Preprocessing}

To ensure consistency across multiple EEG datasets, we follow standard preprocessing procedures, including channel alignment, filtering, resampling, segmentation, and normalization, consistent with the preprocessing pipeline used in LEAD \cite{Wang2025-lead}, as illustrated in the preprocessing block of Figure~1.

\textit{Channel harmonization (19-channel 10--20 montage):}
All EEG datasets were standardized to a common 19-channel configuration based on the international 10--20 system. Datasets originally acquired with 19 electrodes (ADFTD, ADFSU, and AD-Auditory) were used without modification after the common preprocessing steps. The APAVA dataset, originally acquired with 16 electrodes, was mapped to the same 19-channel layout by interpolating the missing channels using spherical spline interpolation \cite{perrin1989spherical}, following the LEAD framework.

First, for a generic electrode position $\mathbf{r}$ on the unit sphere, the scalp potential is modeled as
\[
V(\mathbf{r})=c_0+\sum_{j=1}^{N} c_j\, g_m\!\big(\cos(\mathbf{r},\mathbf{r}_j)\big),
\tag{1}
\]
where $\mathbf{r}$ denotes the electrode position at which the potential is to be estimated, $\mathbf{r}_j$ denotes the position of the $j$-th observed electrode, $c_0$ is a constant term, $c_j$ are interpolation coefficients \cite{perrin1989spherical  }, and $g_m(\cdot)$ is the spherical spline kernel defined as
\[
g_m(x)=\frac{1}{4\pi}\sum_{n=1}^{\infty}\frac{2n+1}{(n(n+1))^m}P_n(x),
\tag{2}
\]
where $P_n(x)$ denotes the Legendre polynomial of degree $n$ \cite{arfken2011mathematical}, defined by
\[
P_n(x)=\frac{1}{2^n n!}\frac{d^n}{dx^n}(x^2-1)^n.
\tag{3}
\]

Second, for the APAVA dataset, which lacks three channels, the positions of the missing electrodes are denoted by $\mathbf{r}_{d_k}$, with $k=1,\dots,K$.
Based on the same kernel, the source--source and destination--source kernel matrices are defined as
\[
G_{ss}[i,j]=g_m\!\big(\cos(\mathbf{r}_i,\mathbf{r}_j)\big), 
\qquad
G_{ds}[k,j]=g_m\!\big(\mathbf{r}_{d_k},\mathbf{r}_j\big),
\tag{5}
\]
where $G_{ss}\in\mathbb{R}^{N\times N}$ is the kernel matrix between the $N$ observed channels, and $G_{ds}\in\mathbb{R}^{M\times N}$ is the kernel matrix between the $M$ missing channels and the $N$ observed channels. Here, $\mathbf{r}_i$ and $\mathbf{r}_j$ denote the positions of the $i$-th and $j$-th observed electrodes, respectively, and $\mathbf{r}_{d_k}$ denotes the position of the $k$-th missing electrode to be reconstructed. The missing-channel signals are then estimated by
\[
\hat{X}
=
\begin{bmatrix}
T_d & G_{ds}
\end{bmatrix}
\begin{bmatrix}
T_s^{T} & 0\\
T_s & G_{ss}
\end{bmatrix}^{-1}
\begin{bmatrix}
0\\
X
\end{bmatrix},
\tag{6}
\]
where $X\in\mathbb{R}^{N\times T}$ is the EEG signal of the $N$ observed channels over $T$ time samples, $\hat{X}\in\mathbb{R}^{M\times T}$ is the reconstructed EEG signal of the $M$ missing channels, and $T_s\in\mathbb{R}^{N\times1}$ and $T_d\in\mathbb{R}^{M\times1}$ are all-ones column vectors. For APAVA, this procedure was used to reconstruct the three missing channels and obtain the full 19-channel representation. Finally, for BrainLat, which was recorded with 128 channels, only electrodes corresponding to the standard 10--20 positions were retained, while the remaining channels were discarded. This standardization ensures a consistent spatial representation across datasets and facilitates reliable cross-dataset analysis. 

\begin{figure}[h] 
    \centering
    \includegraphics[scale=0.35]{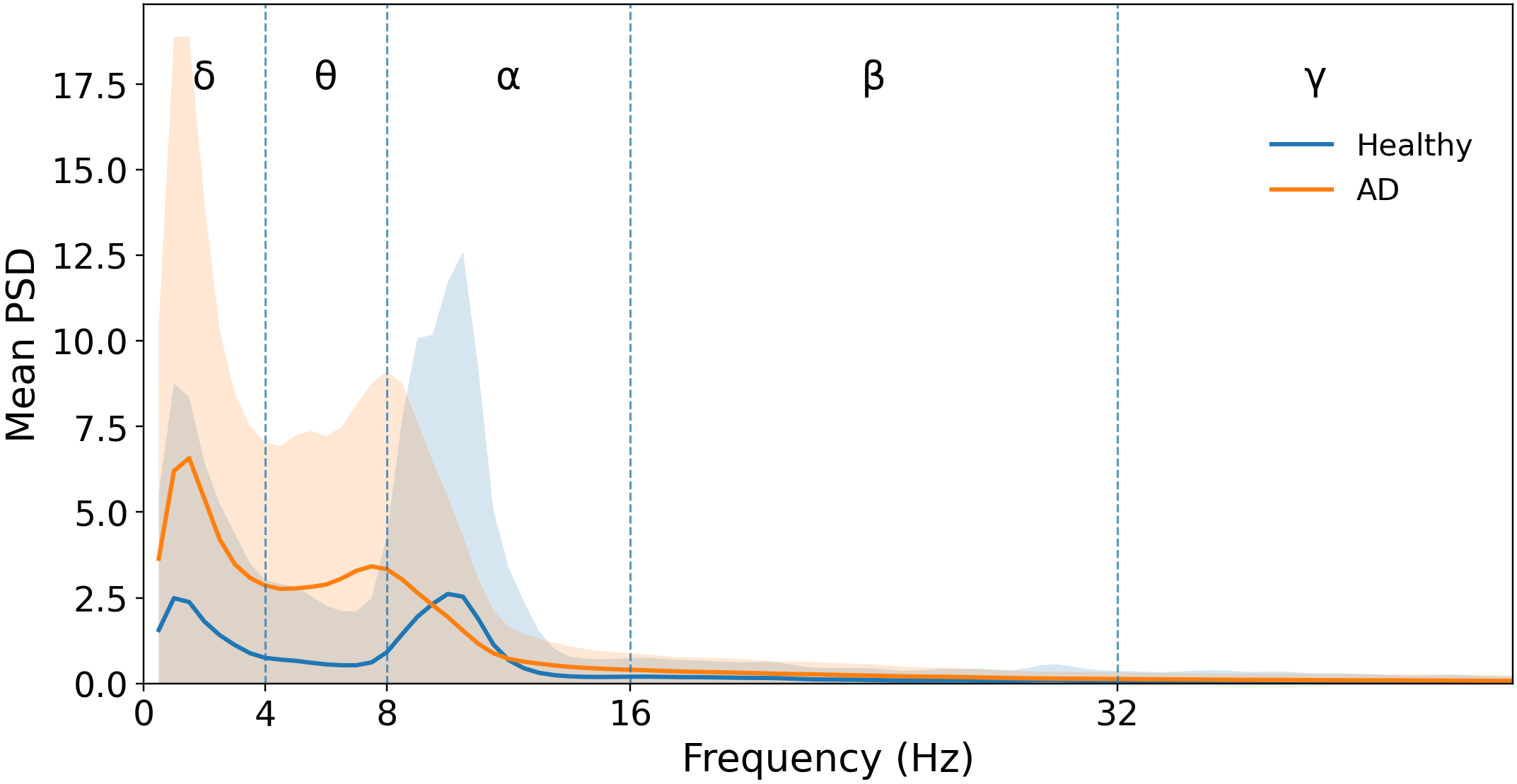}
    \caption{\small \label{fig:raw_data_vis} A comparative frequency-domain analysis of the AD and HC signals across four channels: alpha, beta, gamma, and delta. (Blue line: HC; Red line: AD.)
}
\end{figure}

\textit{Sampling-rate harmonization and filtering:}
Let $x^{\mathrm{raw}}_{c}$ denote the raw EEG signal from channel $c \in {1,\ldots,N_c}$ after harmonization to the standard 10--20 montage ($N_c = 19$). Based on the analysis shown on \ref{fig:raw_data_vis}, we apply a band-pass filter with cut-off frequencies of 0.5--45~Hz to suppress slow drifts and high-frequency noise while preserving canonical EEG rhythms, and the filter is designed according to the sampling rate of each dataset. The filtered signals are then resampled to a common sampling rate of $f_s' = 128$~Hz, yielding the processed signal $\hat{x}_{c}[n]$.


\textit{Wavelet decomposition and frequency-band analysis:}
In this study, the symlet-4 (sym4) mother wavelet is adopted, as it has been reported to yield strong discriminative performance in EEG classification compared with other wavelet families~\cite{Farina2020-oc}. Stationary Wavelet Transform (SWT) is applied independently to each channel using the preprocessed EEG signal $\hat{x}_{c}[n]$ as input, which is set as the level-0 approximation, i.e., $A_{0,c}[n]=\hat{x}_{c}[n]$. At each decomposition level $j$, SWT produces an approximation coefficient sequence $A_{j,c}[n]$ (low-frequency component) and a detail coefficient sequence $D_{j,c}[n]$ (high-frequency component) according to the standard SWT formulation in~\cite{frikha2021source}. For a signal sampled at $f_s'$, the detail coefficients at level $j$ correspond approximately to the frequency interval $D_{j,c}:\left(\frac{f_s'}{2^{j+1}},\,\frac{f_s'}{2^{j}}\right)$, while the approximation at the final level captures $A_{j,c}:\left(0,\,\frac{f_s'}{2^{j+1}}\right)$.With $f_s'=128$~Hz and a four-level decomposition ($j=4$), the resulting SWT sub-bands align with canonical EEG rhythms, and the band-to-subband mapping in Fig~2 is referenced and implemented following the grouping strategy in Aydin et al.~\cite{Aydin2016-wavelet}. Therefore, we construct rhythm-specific time series for each channel by assigning the final-level approximation to the delta band and the detail coefficients to the remaining bands.
\textit{Segmentation and normalization:}
After channel and sampling-rate harmonization and wavelet decomposition, segmentation is applied to each rhythm-specific EEG signal $\hat{x}_{c}^{(r)}[n]$, where $r \in \{\delta,\theta,\alpha,\beta,\gamma\}$. Each EEG signal is segmented into windows of length $L = 128$ samples (corresponding to 1s) with 50\% overlap, resulting in a step size of $L/2$. The $k$-th segment of channel $c$ in rhythm $r$ is obtained by selecting $L$ consecutive samples starting from index $kL/2$. It is defined as
\begin{equation}
\mathbf{\hat{x}}^{(r)}_{c,k}
=
\big[
\hat{x}^{(r)}_{c}[\frac{kL}{2} + 1],\,
\hat{x}^{(r)}_{c}[\frac{kL}{2} + 2],\,
\dots,\,
\hat{x}^{(r)}_{c}[\frac{kL}{2} + L]
\big]
\end{equation}

Each segment is subsequently normalized using per-channel z-score normalization computed within the segment, following the procedure used in LEAD. After normalization, the resulting signal is denoted as $x^{(r)}_{c,k}$.

\subsection{DeepTokenEEG Architecture}

\begin{figure}[h] 
    \centering
    \includegraphics[scale=0.25]{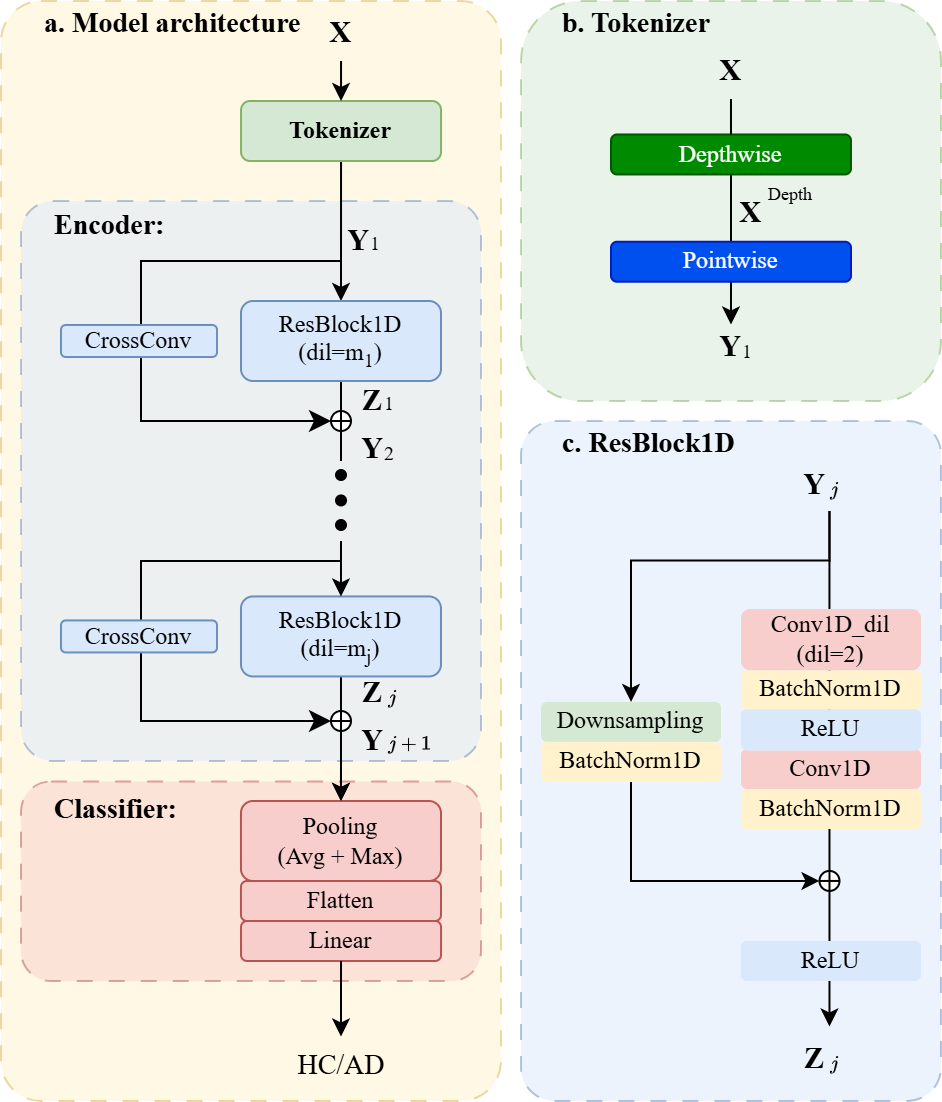}
    \caption{\small \label{fig:model_arch}Overall architecture of the proposed DeepTokenEEG model. 
\textbf{(a)} Overall framework comprising the Tokenizer, Encoder (featuring three ResBlock1D stages with parallel CrossConv branches), and Classifier. \textbf{(b)} Detailed view of the Tokenizer utilizing Depthwise and Pointwise convolutions. \textbf{(c)} Structure of the General ResBlock1D employing progressively dilated convolutions. }
\end{figure}

The proposed model is divided into three main components: Tokenizer, Encoder, and Classifier, as shown in Fig. \ref{fig:model_arch}.

\subsubsection{Tokenizer}

Given the preprocessed EEG dataset, the preprocessing stage outputs rhythm-specific segments denoted by $\mathbf{x}^{(r)}_{c,k}$, where $r \in \{\delta,\theta,\alpha,\beta,\gamma\}$ indexes the frequency band and $k$ indexes the segment. For model input, the first segment from the alpha band is selected and denoted as $x^{(\alpha)}_{c,1}$. For simplicity, this segment is written as the discrete time-series vector
\[
\mathbf{x}_c = [x_c(1), x_c(2), \dots, x_c(L)],
\]
where $x_c(\ell)$ is the EEG amplitude at time step $\ell$ within the segment, with $1 \le \ell \le L$.

\begin{figure*}[h] 
    \centering
    \includegraphics[scale=0.46]{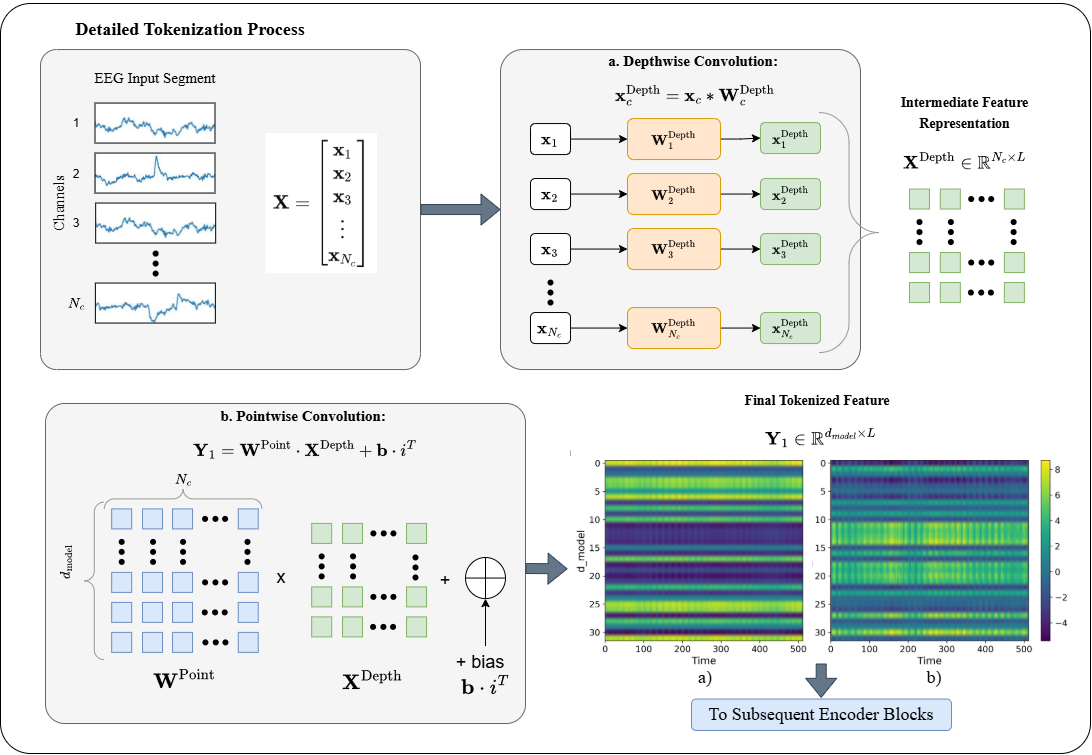}
    \caption{\small\label{fig:tokenizer_detailed_process}
    Detailed illustration of the tokenizer module used in DeepTokenEEG. An input EEG segment $\mathbf{X}\in\mathbb{R}^{N_c\times L}$ is first processed by a channel-wise depthwise convolution to extract local temporal patterns independently from each electrode, producing an intermediate representation $\mathbf{X}^{\mathrm{Depth}}\in\mathbb{R}^{N_c\times L}$. A subsequent pointwise convolution then linearly combines information across channels and projects it into a latent embedding space, yielding the tokenized feature map $\mathbf{Y}_1\in\mathbb{R}^{d_{\mathrm{model}}\times L}$. The two heatmaps at the bottom present representative tokenizer outputs for two EEG samples, with panel (a) showing a healthy control (HC) subject and panel (b) showing an Alzheimer’s disease (AD) subject. The horizontal axis denotes time, while the vertical axis represents learned latent feature dimensions.
    }
\end{figure*}

A single EEG segment is then formed by stacking all channel vectors into the matrix
\begin{equation}
    \mathbf{X} = [\mathbf{x}_1 \quad \mathbf{x}_2 \quad \cdots \quad \mathbf{x}_{N_c}]^T \in \mathbb{R}^{N_c \times L}.
\end{equation}
This matrix serves as the fundamental input unit to the tokenizer block.

To efficiently capture both temporal structure within each channel and spatial dependencies across channels, we employ a depthwise separable tokenizer. The module consists of two consecutive stages: depthwise convolution and pointwise convolution. This design allows the tokenizer to preserve temporal resolution while transforming the raw multi-channel EEG into a compact latent representation suitable for subsequent encoder blocks. More specifically, for an input segment with temporal length $L$ and $N_c$ channels, the tokenizer produces an output representation of size $\mathbf{Y}_1 \in \mathbb{R}^{d_{\text{model}}\times L}$, where each row corresponds to a learned latent feature dimension and each column corresponds to a temporal sample. Therefore, the temporal dynamics of the EEG signal are retained, while the original channel space is mapped into a learned embedding space.

\paragraph{Depthwise Convolution}

In the first stage, each EEG channel is processed independently using a 1D depthwise convolution layer. Let $\mathbf{W}^{\text{Depth}}$ denote the collection of depthwise convolution weights, composed of learnable filter kernels $\mathbf{W}^{\text{Depth}}_c$ assigned to the $c$-th channel. This operation enables the model to extract local temporal structure from each channel without mixing information across electrodes. The depthwise convolution for the $c$-th channel is given by
\begin{equation}
\mathbf{x}^{\text{Depth}}_{c} = \mathbf{x}_{c} * \mathbf{W}_{c}^{\text{Depth}},
\end{equation}
where $*$ denotes the convolution operator. Here, $\mathbf{x}^{\text{Depth}}_{c}$ is the transformed temporal representation of channel $c$, with length $L$. By stacking these filtered signals for all $N_c$ channels, we obtain the intermediate feature matrix $\mathbf{X}^{\text{Depth}} \in \mathbb{R}^{N_c \times L}$
which preserves the input dimensions while encoding channel-specific temporal features.

This stage is particularly important for EEG modeling because it allows each electrode to retain its own local temporal dynamics before any cross-channel mixing is introduced. As a result, short-range rhythmic patterns, local oscillatory variations, and transient waveform characteristics can be captured separately for each channel.

\paragraph{Pointwise Convolution}
Following the depthwise stage, a pointwise convolution is applied to mix channel-wise ìnormation. Functioning as a linear projection, this operation transforms the channel features into a high-dimensional latent space ($d_\text{model}$) by mixing information across all electrodes. 

Let $\mathbf{W}^{\text{Point}} \in \mathbb{R}^{d_\text{model} \times N_c}$ denote the trainable weight matrix and ${b} \in \mathbb{R}^{d_\text{model}}$ be the trainable bias vector. The projection for the input segment is computed as:
\begin{equation}
    \mathbf{Y}_1 = \mathbf{W}^{\text{Point}}\mathbf{X}^{\text{Depth}} + \mathbf{b}\mathbf{i}^T,
\end{equation}
where $\mathbf{i}$ is an all-ones vector of size $L$, used to broadcast the bias along the temporal dimension. The inclusion of $\mathbf{b}$ provides an affine shift, allowing the model to represent latent feature distributions that are not necessarily centered at the origin. The resulting output $\mathbf{Y}_1 \in \mathbb{R}^{d_\text{model} \times L}$ serves as the tokenized representation and is passed to the subsequent encoder blocks.

From a representation-learning perspective, this pointwise projection is the stage at which cross-channel EEG information is fused into a shared latent space. Importantly, the vertical axis of $\mathbf{Y}_1$ no longer corresponds to the original EEG electrodes. Instead, it indexes learned latent feature channels that summarize information aggregated from multiple electrodes and their local temporal neighborhoods. Likewise, the values in $\mathbf{Y}_1$ are not physical EEG amplitudes (e.g., in $\mu$V), but dimensionless latent activations produced by the tokenizer.

The tokenizer is designed to provide an informative first-stage representation prior to deep encoding. In practice, the resulting latent feature maps tend to exhibit structured temporal organization. EEG segments with relatively stable temporal patterns produce smoother and more coherent activation trajectories across latent dimensions, whereas EEG segments from Alzheimer’s disease (AD) subjects tend to show less stable temporal organization, leading to stronger temporal fluctuations and more fragmented activation patterns in the latent space. This behavior is consistent with the intended role of the tokenizer: to preserve relevant temporal structure, aggregate spatial information across electrodes, and expose discriminative signal characteristics in a form that can be more effectively processed by the downstream encoder.

In this way, the tokenizer performs more than a simple dimensional projection. It acts as an initial feature extractor that reshapes raw EEG segments into temporally aligned latent tokens, establishing the representational basis for subsequent hierarchical modeling.

\subsubsection{ Encoder}

The encoder block serves as the principal feature extraction component of the model, receiving the output $\mathbf{Y}_1$ from the tokenizer block, as illustrated in Fig.~\ref{fig:model_arch}a. 
Its architecture consists of $j$ cascaded stages, where $1 \le j \le 5$, and the exact number of stages is not fixed a priori but determined empirically through experimental evaluation. 
The maximum number of stages is constrained to five in order to balance computational cost, model accuracy, and the adverse effects of excessive dilation, which may lead to feature sparsification and loss of critical local information. Each stage $j$ is composed of two parallel processing branches, those are 1D Residual Block (ResBlock1D) and {Cross Convolution Branch (CrossConv).

\textbf{ResBlock1D branch} is the main processing branch, responsible for generating the intermediate representation denoted as $\mathbf{Z}_j$. Each ResBlock1D adopts a bottleneck architecture consisting of two sequential convolutional layers. The first layer employs a dilated convolution with dilation rate $m_j$, followed by Batch Normalization and a ReLU activation function. In this work, the dilation rate is fixed to $m_j = 2$ for all ResBlock1D stages. The second layer applies a standard convolution followed by Batch Normalization. A shortcut connection is implemented using a Conv1D downsampling layer followed by Batch Normalization, allowing the block input to be added directly to the output of the second Batch Normalization layer. A final ReLU activation function is applied after this element-wise addition to produce $\mathbf{Z}_j \in \mathbb{R}^{d_\text{model} \times L}$. This design expands the effective receptive field of the model while maintaining a reasonable balance between global context modeling and local feature preservation. Moreover, employing dilation within each ResBlock1D contributes to reducing computational cost compared to stacking additional convolutional layers. The optimal number of ResBlock1D stages and the corresponding dilation configuration are determined through ablation studies to quantify the contribution of each stage to the overall performance.

\textbf{CrossConv branch} acts as an external shortcut path, employing a trainable $1 \times 1$ kernel 1D convolution layer to mitigate vanishing gradient issues and facilitate feature reuse. A downsampling operation is also incorporated in this branch to ensure dimensional consistency between $\mathbf{Y}_j$ and $\mathbf{Z}_j$ after the application of dilation.

The processing flow within each stage $j$ consists of two steps. 
First, the ResBlock1D produces an intermediate output:
\begin{equation}
    \mathbf{Z}_j = \text{ResBlock1D}_j(\mathbf{Y}_j),
\end{equation}
where $\mathbf{Y}_j \in \mathbb{R}^{d_\text{model} \times L}$ denotes the input feature representation at stage $j$. 
Subsequently, this intermediate output is combined with the output of the CrossConv branch as follows:
\begin{equation}
    \mathbf{Y}_{j+1} = \mathbf{Z}_j + \mathbf{a}_j \cdot \mathbf{Y}_j,
\end{equation}
where $\mathbf{Y}_{j+1} \in \mathbb{R}^{d_\text{model} \times L}$ represents the output feature map of stage $j$.

After passing through all encoder stages, the final encoder output is obtained and forwarded to the subsequent classifier block.

\subsubsection{Classifier}
To aggregate temporal information from the encoder output, both adaptive average pooling and adaptive max pooling are applied along the temporal dimension. 
These operations compress the variable-length temporal features into fixed-size representations while preserving complementary statistical information. 
The pooled features are then concatenated and flattened to form a unified feature vector for each input sample.

The resulting vector is fed into a classification head consisting of layer normalization followed by a fully connected layer, which maps the aggregated features to the target classes. 
The model is trained using the cross-entropy loss for classification tasks. 

\section{Experiment}
\label{Experiment}

\subsection{Data Splitting and Baseline:}


To ensure a reliable and robust evaluation, we employed a subject-independent 5-fold cross-validation strategy. In each fold, the subjects (grouped into HC and AD) were randomly partitioned into training and test subsets, effectively mitigating data leakage. 


To evaluate the performance of DeepTokenEEG, four distinct comparisons were conducted. First, A systematic ablation study was performed to determine the optimal model configuration, including evaluating the impact of varying the number of encoder blocks to maximize feature extraction capability. Second, the classification accuracy of the proposed model was assessed and compared against existing baselines. Subsequently, we investigated the importance of different sub-bands of electroencephalogram (alpha, beta, delta, theta, gamma) in classifying AD and HC compared to the full-band signal. This analysis identified biomarkers in the frequency domain that have the highest discriminative potential for detecting Alzheimer's disease. Finally, the number of parameters and computational complexity (GFLOPs) of each method were also included in the comparison to emphasize the model's performance and efficiency. For this purpose, we compared our method with 10 baselines, consisting of 5 supervised learning methods, 3 self-supervised learning methods, and 2 large-scale EEG foundation models, along with three variants of the contrastive learning approach LEAD: LEAD-Vanilla, LEAD-Sub, and LEAD-Base. The selected baselines are either state-of-the-art methods or have demonstrated strong performance in EEG or time series classification tasks. Specifically, the 5 supervised learning methods include TCN \cite{Bairagi2018-dy}, vanilla Transformer \cite{vaswani2017attention}, Conformer \cite{Song2023}, TimesNet \cite{wu2023timesnet}, and Medformer \cite{10.1145/3708036.3708089}. The 3 self-supervised learning methods are TS2Vec \cite{yue2022ts2vec}, BIOT \cite{yang2023biot}, and EEG2Rep \cite{MohammadiFoumani2024}. The 2 large-scale EEG foundation models are LaBraM \cite{Jiang2024LargeBM} and EEGPT \cite{He2024}.

\subsection{Training Setup}

All models were trained from scratch using identical training configurations to ensure a fair comparison across different architectures and data settings. 
Experiments were conducted independently on three dataset settings, including ADFTD, BrainLat, and the Combined Dataset, following a subject-independent 5-fold cross-validation protocol.

EEG signals were segmented into fixed-length sequences of 128 samples, corresponding to 1-second windows at a sampling rate of 128 Hz. 
For each fold, the models were trained for 100 epochs with a batch size of 128 on a single NVIDIA GeForce RTX 3090 Ti GPU. 
The Adam optimizer was employed with an initial learning rate of $1 \times 10^{-4}$, and a dropout rate of 0.1 was applied to mitigate overfitting.

\begin{table}[t]
\centering
\caption{Training configuration used for all experiments.}
\label{tab:training_setup}
\begin{tabular}{ll}
\toprule
\textbf{Parameter} & \textbf{Value} \\
\midrule
Cross-validation protocol & Subject-independent 5-fold CV \\
Input sequence length & 128 samples (1 s at 128 Hz) \\
Batch size & 128 \\
Training epochs & 100 \\
Optimizer & Adam \\
Initial learning rate & $1 \times 10^{-4}$ \\
Dropout rate & 0.1 \\
Loss function & Cross-entropy \\
Hardware & NVIDIA GeForce RTX 3090 Ti GPU \\
\bottomrule
\end{tabular}
\end{table}

To evaluate the effect of different frequency-band representations, the proposed DeepTokenEEG framework was trained using EEG signals filtered into distinct frequency bands, including full-band and canonical sub-bands. 
The final performance metrics are reported as the mean $\pm$ standard deviation across all five folds to ensure statistical reliability.

We evaluated all baseline methods and the proposed model using a consistent set of metrics to ensure fair and reliable comparisons. Model performance was assessed at both the segment level and the subject level. The primary evaluation metrics included F1-score (F1) and Accuracy (Acc), which jointly reflect the general correctness of the model and its ability to distinguish Alzheimer's patients from healthy controls.

In sample-level evaluation, each EEG segment (1-second window) is treated as an independent instance, and performance is calculated over all segments. This strategy is commonly adopted in previous works, where feature extraction and classification are performed per segment. In contrast, subject-level evaluation aggregates predictions across all segments belonging to a subject to generate a final diagnostic decision.

\begin{equation}
    \text{Precision} = \frac{\text{TP}}{\text{TP} + \text{FP}}
\end{equation}

\begin{equation}
    \text{Recall} = \frac{\text{TP}}{\text{TP} + \text{FN}}
\end{equation}

\begin{equation}
    \text{F1} = \frac{2 \cdot \text{Precision} \cdot \text{Recall}}{\text{Precision} + \text{Recall}}
\end{equation}

\begin{equation}
    \text{Accuracy} = \frac{\text{TP} + \text{TN}}{\text{TP} + \text{TN} + \text{FP} + \text{FN}}
\end{equation}

In this context, true positives (TP) and true negatives (TN) denote AD and HC subjects correctly classified, respectively, whereas false positives (FP) and false negatives (FN) correspond to HC subjects misclassified as AD and AD subjects misclassified as HC. Accuracy reflects the overall proportion of correct classifications, precision indicates the proportion of predicted AD cases that are truly AD, recall measures the proportion of AD subjects correctly identified.

To ensure statistical robustness, each experiment was repeated five times with different random data splits. The final results are reported as the mean mean $\pm$ standard deviation for these runs.


\section{Results}
\label{Results}

Experimental results are first analyzed to determine the optimal number of encoder stages and dilation settings. 
The selected architecture is subsequently used for in-depth performance evaluation and comparative analysis against competing methods.

\subsection{Encoder Stage Ablation}

To further isolate the effect of encoder depth, Table~\ref{tab:ablation_nblocks_dil2} reports an ablation study in which the dilation rate is fixed to $m_j=2$ for all stages.
Under this controlled setting, the influence of the number of encoder stages can be more clearly assessed.
The results demonstrate that an encoder with three stages consistently achieves superior performance at both the segment and subject levels.
In particular, the three-stage configuration yields perfect subject-level accuracy and F1-score, while also providing strong segment-level performance.

Although deeper encoders with four or five stages remain competitive, their performance does not improve proportionally to the increased model complexity.
Therefore, considering both classification accuracy and computational efficiency, an encoder composed of three ResBlock1D stages with a uniform dilation of $m_j=2$ is selected as the optimal architecture.
This configuration offers a favorable balance between representational capacity, training stability, and computational cost, and is adopted in all subsequent experiments and comparisons.

\begin{table}[h]
\centering
\caption{Ablation study evaluating the effect of progressively adding ResBlock1D stages using the best dilation setting ($m_j=2$ for all stages). Results are reported as mean$\pm$std (\%) and input is $\alpha-\text{band}$ of BrainLat-testset.}
\label{tab:ablation_nblocks_dil2}
\resizebox{0.48\textwidth}{!}{%
\begin{tabular}{rcccc}
\toprule
\textbf{Stage $j$} & \textbf{subj\_f1 (\%)} & \textbf{subj\_acc (\%)} & \textbf{seg\_f1 (\%)} & \textbf{seg\_acc (\%)} \\
\midrule
1 ($m_j=2$) & 92.69$\pm$7.11  & 92.79$\pm$6.95  & 90.70$\pm$0.53 & 90.71$\pm$0.53 \\
2 ($m_j=2$) & 92.69$\pm$7.11  & 92.79$\pm$6.95  & 91.19$\pm$0.52 & 91.20$\pm$0.52 \\
\textbf{3 ($m_j=2$)} & \textbf{100.00$\pm$0.00} & \textbf{100.00$\pm$0.00} & \textbf{96.68$\pm$0.34} & \textbf{96.68$\pm$0.34} \\
4 ($m_j=2$) & 92.69$\pm$7.11  & 92.79$\pm$6.95  & 92.88$\pm$0.48 & 92.88$\pm$0.49 \\
\textbf{5 ($m_j=2$)} & \textbf{92.69$\pm$7.11}  & \textbf{92.79$\pm$6.95}  & \textbf{96.90$\pm$0.33} & \textbf{96.89$\pm$0.33} \\
\bottomrule
\end{tabular}
}
\end{table}

\subsection{Classification Performance}

Table~\ref{tab:adftd_brainlat_combined_fullband} summarizes the classification results of DeepTokenEEG and competing baselines on the ADFTD, BrainLat, and Combined datasets using full-band EEG signals.
Overall, DeepTokenEEG consistently delivers the strongest or near-strongest performance across both segment-level classification and subject-level detection, while using only 0.29M parameters.
This result indicates that the proposed lightweight tokenizer-encoder design is highly effective at extracting discriminative EEG representations without relying on large-scale architectures.

On the ADFTD dataset, DeepTokenEEG achieves the best performance at both evaluation levels, reaching 85.14\% segment-level accuracy and 81.92\% segment-level F1-score, as well as 87.12\% subject-level accuracy and 86.92\% subject-level F1-score.
These results outperform all compared baselines, including stronger recent methods such as LEAD-Sup, BIOT, and TS2Vec.
This suggests that the proposed architecture can effectively capture disease-related temporal patterns in relatively cleaner and more homogeneous EEG recordings.

A similar trend is observed on the BrainLat dataset.
DeepTokenEEG obtains the best segment-level performance, with 85.71\% accuracy and 85.42\% F1-score, and also achieves the highest subject-level accuracy and F1-score among all methods.
Compared with conventional temporal models such as TCN and Transformer, as well as recent EEG foundation-inspired models such as BIOT, LaBraM, and EEGPT, the proposed method shows substantially stronger generalization.
These findings indicate that the tokenizer is able to preserve informative temporal dynamics and inter-channel structure, which remain discriminative even under cross-dataset variations.

\paragraph{Segment-Level vs. Subject-Level Performance}
Across all methods, subject-level results are generally higher and more stable than segment-level results.
This behavior is expected because subject-level prediction aggregates evidence from multiple EEG segments, thereby reducing the impact of noisy or ambiguous individual segments.
For DeepTokenEEG, this aggregation further improves robustness, leading to consistently strong subject-level detection performance across all datasets.
The improvement from segment-level to subject-level evaluation also confirms that the learned segment representations are sufficiently informative and complementary when combined at the subject level.

\paragraph{Performance on the Combined Dataset}
Although DeepTokenEEG still achieves the best overall performance on the Combined dataset, the classification accuracy is lower than that obtained on ADFTD or BrainLat individually.
This degradation is expected because the Combined dataset merges several independent EEG cohorts collected under different acquisition settings, clinical protocols, subject populations, and data qualities.
As a result, the model must handle substantially greater inter-dataset heterogeneity, including differences in noise level, recording devices, electrode placement conventions, preprocessing pipelines, class distributions, and disease severity.
Such variability increases the difficulty of learning a unified decision boundary and naturally reduces classification performance for all methods.

Another plausible explanation is that some constituent datasets in the Combined setting may contain lower-quality or more challenging EEG samples, which can pull down the overall performance when pooled together.
In other words, the Combined dataset is not simply larger, but also harder, because it introduces additional domain shift and label uncertainty across cohorts.
Therefore, the lower accuracy on the Combined dataset should not be interpreted as a weakness of the proposed method, but rather as evidence of the increased complexity of large-scale multi-source EEG classification.
Importantly, despite this more challenging setting, DeepTokenEEG still maintains the best segment-level accuracy and F1-score, as well as the best subject-level accuracy and F1-score, demonstrating strong robustness and generalization ability.

\paragraph{Why DeepTokenEEG Performs Better}
The superior performance of DeepTokenEEG can be attributed to three main factors.
First, the tokenizer provides an efficient front-end representation that jointly captures temporal dynamics and cross-channel EEG structure at an early stage.
Second, the compact encoder design avoids unnecessary model complexity, which helps reduce overfitting on relatively limited EEG datasets.
Third, the proposed architecture appears to preserve disease-relevant temporal irregularities while suppressing redundant background activity, resulting in more separable representations for HC and AD subjects.
These properties explain why DeepTokenEEG is able to outperform larger and computationally heavier baselines, even though it uses substantially fewer parameters.

\begin{table*}[ht!]
\centering
\caption{Performance comparison of DeepTokenEEG and baseline methods on the ADFTD, BrainLat, and Combined dataset (ADFTD, BrainLat, AD-Auditory, ADFSU, APAVA) at the segment and subject levels, when trained and evaluated using full-band EEG signals.}
\label{tab:adftd_brainlat_combined_fullband}
\small 
\begin{tabularx}{\textwidth}{>{\raggedright\arraybackslash}X|cc|cc|cc}
\toprule
\textbf{Datasets} 
& \multicolumn{2}{c|}{\textbf{ADFTD}} 
& \multicolumn{2}{c|}{\textbf{BrainLat}} 
& \multicolumn{2}{c}{\textbf{Combined}} \\
\midrule
\textbf{Method (Params)} 
& \textbf{Accuracy (\%)} & \textbf{F1-Score (\%)} 
& \textbf{Accuracy (\%)} & \textbf{F1-Score (\%)}
& \textbf{Accuracy (\%)} & \textbf{F1-Score (\%)} \\
\midrule

\multicolumn{7}{c}{\textbf{Segment-Level Classification Results}} \\
\midrule
\textbf{TCN (1.02M)} \cite{Bairagi2018-dy} 
& 75.01$\pm$0.95 & 74.90$\pm$0.90 
& 59.23$\pm$0.97 & 59.21$\pm$0.98 
& 60.33$\pm$2.71 & 62.24$\pm$3.07 \\

\textbf{Transformer (0.83M)} \cite{vaswani2017attention} 
& 67.89$\pm$1.24 & 67.29$\pm$2.53 
& 59.21$\pm$1.05 & 58.60$\pm$0.83 
& 55.99$\pm$4.43 & 50.60$\pm$5.85 \\

\textbf{Conformer (1.14M)} \cite{Song2023} 
& 75.04$\pm$1.56 & 74.80$\pm$1.66 
& 62.45$\pm$1.09 & 60.12$\pm$1.66 
& 57.41$\pm$7.59 & 60.07$\pm$6.68 \\

\textbf{TimesNet (2.35M)} \cite{wu2023timesnet} 
& 65.06$\pm$1.59 & 74.42$\pm$2.07 
& 59.63$\pm$1.56 & 59.22$\pm$1.45 
& 62.31$\pm$4.25 & 67.31$\pm$6.16 \\

\textbf{Medformer (4.42M)} \cite{10.1145/3708036.3708089} 
& 73.88$\pm$1.23 & 73.77$\pm$1.19 
& 60.15$\pm$0.79 & 59.86$\pm$0.86 
& 64.06$\pm$2.81 & 69.09$\pm$4.39 \\

\textbf{TS2Vec (2.58M)} \cite{yue2022ts2vec} 
& 71.81$\pm$0.84 & 71.73$\pm$0.89 
& 67.94$\pm$1.10 & 67.94$\pm$1.10 
& 64.56$\pm$2.42 & 67.29$\pm$4.46 \\

\textbf{BIOT (4.16M)} \cite{yang2023biot} 
& 78.63$\pm$0.80 & 77.43$\pm$0.89 
& 61.51$\pm$1.50 & 61.51$\pm$1.50 
& 59.16$\pm$4.00 & 70.36$\pm$6.93 \\

\textbf{EEG2Rep (5.33M)} \cite{MohammadiFoumani2024} 
& 70.62$\pm$1.31 & 70.60$\pm$1.32 
& 68.02$\pm$3.86 & 67.60$\pm$3.87 
& 64.69$\pm$4.37 & 65.07$\pm$8.86 \\

\textbf{LaBraM (9.62M)} \cite{Jiang2024LargeBM} 
& 55.07$\pm$0.90 & 71.03$\pm$0.80 
& 48.24$\pm$0.70 & 48.24$\pm$0.70 
& 63.40$\pm$3.79 & 67.94$\pm$5.09 \\

\textbf{EEGPT (25.5M)} \cite{He2024} 
& 54.09$\pm$0.70 & 71.00$\pm$0.00 
& 49.09$\pm$0.00 & 65.85$\pm$0.00 
& 65.11$\pm$2.77 & 67.36$\pm$5.52 \\

\textbf{LEAD-Sup (3.21M)} 
\cite{Wang2025-lead}
& 80.84$\pm$0.84 & 80.68$\pm$0.79 
& 70.36$\pm$0.67 & 70.30$\pm$0.62 
& 64.75$\pm$3.09 & 69.18$\pm$5.59 \\

\textbf{DeepTokenEEG (0.29M)} 
& \textbf{85.14$\pm$3.65} & \textbf{81.92$\pm$1.19} 
& \textbf{82.12$\pm$2.52 } & \textbf{81.42$\pm$2.76} 
& \textbf{68.03$\pm$3.85} & \textbf{73.96$\pm$3.83} \\

\midrule
\multicolumn{7}{c}{\textbf{Subject-Level Detection Results}} \\
\midrule

\textbf{TCN (1.02M)} \cite{Bairagi2018-dy} 
& 81.43$\pm$3.50 & 81.36$\pm$3.55 
& 75.71$\pm$7.28 & 75.57$\pm$7.24 
& 58.36$\pm$4.03 & 65.14$\pm$5.81 \\

\textbf{Transformer (0.83M)} \cite{vaswani2017attention} 
& 74.29$\pm$3.50 & 74.00$\pm$3.65 
& 74.29$\pm$5.71 & 72.82$\pm$6.12 
& 67.95$\pm$2.03 & 73.47$\pm$2.86 \\

\textbf{Conformer (1.14M)} \cite{Song2023} 
& 85.71$\pm$4.35 & 76.86$\pm$5.95 
& 68.57$\pm$1.61 & 40.15$\pm$15.55 
& 67.56$\pm$3.33 & 75.30$\pm$2.82 \\

\textbf{TimesNet (2.35M)} \cite{wu2023timesnet} 
& 81.43$\pm$3.50 & 81.05$\pm$3.18 
& 77.14$\pm$5.35 & 76.23$\pm$5.09 
& 70.26$\pm$7.61 & 78.31$\pm$5.66 \\

\textbf{Medformer (4.42M)} \cite{10.1145/3708036.3708089} 
& 78.57$\pm$0.00 & 78.46$\pm$0.00 
& 74.29$\pm$5.71 & 73.51$\pm$5.97 
& 75.23$\pm$7.33 & 83.32$\pm$6.18 \\

\textbf{TS2Vec (2.58M)} \cite{yue2022ts2vec} 
& 78.57$\pm$0.00 & 78.46$\pm$0.00 
& 84.29$\pm$2.86 & 84.29$\pm$2.86 
& 77.18$\pm$6.34 & 83.46$\pm$6.08 \\

\textbf{BIOT (4.16M)} \cite{yang2023biot} 
& 85.71$\pm$0.00 & 84.44$\pm$0.00 
& 64.29$\pm$6.39 & 63.38$\pm$6.28 
& 74.44$\pm$3.81 & 84.44$\pm$2.88 \\

\textbf{EEG2Rep (5.33M)} \cite{MohammadiFoumani2024} 
& 75.71$\pm$5.71 & 75.19$\pm$5.34 
& 78.57$\pm$5.71 & 78.57$\pm$5.71 
& 70.68$\pm$6.49 & 76.80$\pm$7.35 \\

\textbf{LaBraM (9.62M)} \cite{Jiang2024LargeBM} 
& 57.14$\pm$0.00 & 72.73$\pm$0.00 
& 66.67$\pm$0.00 & 66.67$\pm$0.00 
& 76.00$\pm$5.80 & 83.19$\pm$5.09 \\

\textbf{EEGPT (25.5M)} \cite{He2024} 
& 57.14$\pm$0.00 & 72.73$\pm$0.00 
& 50.00$\pm$0.00 & 66.67$\pm$0.00 
& 77.12$\pm$3.00 & 83.40$\pm$3.64 \\

\textbf{LEAD-Sup (3.21M)} 
\cite{Wang2025-lead}
& 81.43$\pm$2.86 & 81.34$\pm$2.81 
& 78.57$\pm$1.54 & 78.57$\pm$1.54 
& 76.31$\pm$4.25 & 83.43$\pm$2.77 \\

\textbf{DeepTokenEEG (0.29M)} 
& \textbf{87.12$\pm$3.45} & \textbf{86.92$\pm$3.16} 
& \textbf{ 85.71$\pm$3.50} & \textbf{85.42$\pm$4.00} 
& \textbf{78.23$\pm$3.52} & \textbf{83.70$\pm$2.95} \\
\bottomrule
\end{tabularx}
\end{table*}

\subsection{Frequency Specific Performance}

To evaluate the contribution of different frequency bands to the model's performance, we conducted analyzes on each individual frequency band (alpha, beta, delta, gamma, and theta) and the full frequency bands. The results summarized in \ref{tab:deeptokeneeg_band_summary} shows that performance metrics at the subject-level consistently outperformed those at the segment-level across all frequency bands. Specifically, the accuracy at the subject-level ranged from $78.23\%$ to $87.44\%$, while the accuracy at the segment-level remained between $68.03\%$ and $73.26\%$. This trend is also reflected in the F1-scores. This difference indicates that the decision-aggregation mechanism by majority vote in multiple segments provides a more stable and reliable diagnostic outcome for individual patients.

The low-$\gamma$ band determined the most discriminative feature, achieving a peak subject-level accuracy of $87.44\%$ with an F1-score of $91.55\%$. Traditional low-frequency bands, often known for the "slowing"characteristic of EEG in Alzheimer's disease, also show strong diagnostic potential. Their subject-level accuracy yield over $84\%$. This result highlights a critical insight: decomposing the raw EEG signal into specific sub-bands via Stationary Wavelet Transform (SWT) allows the model to isolate relevant features and reduce spectral interference. By focusing on band-specific biological deviations, the model avoids the "noise-masking" effect inherent in the complex, raw full-band signal. 
\begin{table}[h]
\centering
\caption{Performance of DeepTokenEEG variants trained and evaluated on the Combined dataset using SWT-based frequency-band inputs: $\delta$ (0--4 Hz), $\theta$ (4--8 Hz), $\alpha$ (8--16 Hz), $\beta$ (16--32 Hz), and low-$\gamma$ (32--45 Hz).}
\label{tab:deeptokeneeg_band_summary}

\small
\setlength{\tabcolsep}{3.5pt}      
\renewcommand{\arraystretch}{1.10} 

\resizebox{\columnwidth}{!}{%
\begin{tabular}{l|cc|cc}
\toprule
\textbf{Scenario} 
& \multicolumn{2}{c|}{\textbf{Segment-Level}} 
& \multicolumn{2}{c}{\textbf{Subject-Level}} \\
\midrule
\textbf{Band} & \textbf{Accuracy (\%)} & \textbf{F1-Score (\%)} & \textbf{Accuracy (\%)} & \textbf{F1-Score (\%)} \\
\midrule
\textbf{$\alpha$} & 69.82 $\pm$ 4.12 & 75.46 $\pm$ 3.07 & 84.72 $\pm$ 4.43 & 90.08 $\pm$ 2.81 \\
\textbf{$\beta$}  & 71.95 $\pm$ 1.88 & 76.37 $\pm$ 2.55 & 84.72 $\pm$ 1.84 & 89.66 $\pm$ 1.34 \\
\textbf{$\delta$} & 69.03 $\pm$ 4.16 & 74.75 $\pm$ 2.98 & 86.25 $\pm$ 3.33 & 91.24 $\pm$ 1.96 \\
\textbf{$\gamma$} & 73.26 $\pm$ 5.95 & 76.39 $\pm$ 5.67 & 87.44 $\pm$ 4.53 & 91.55 $\pm$ 3.02 \\
\textbf{$\theta$} & 69.00 $\pm$ 4.57 & 71.73 $\pm$ 4.78 & 85.07 $\pm$ 3.43 & 89.75 $\pm$ 2.15 \\
\textbf{Fullband} & 68.03 $\pm$ 3.85 & 73.96 $\pm$ 3.83 & 78.23 $\pm$ 3.52 & 83.70 $\pm$ 2.95 \\
\bottomrule
\end{tabular}%
}
\end{table}

\subsection{Computational Efficiency}



\begin{figure}[h]
\centering
\includegraphics[scale=0.36]{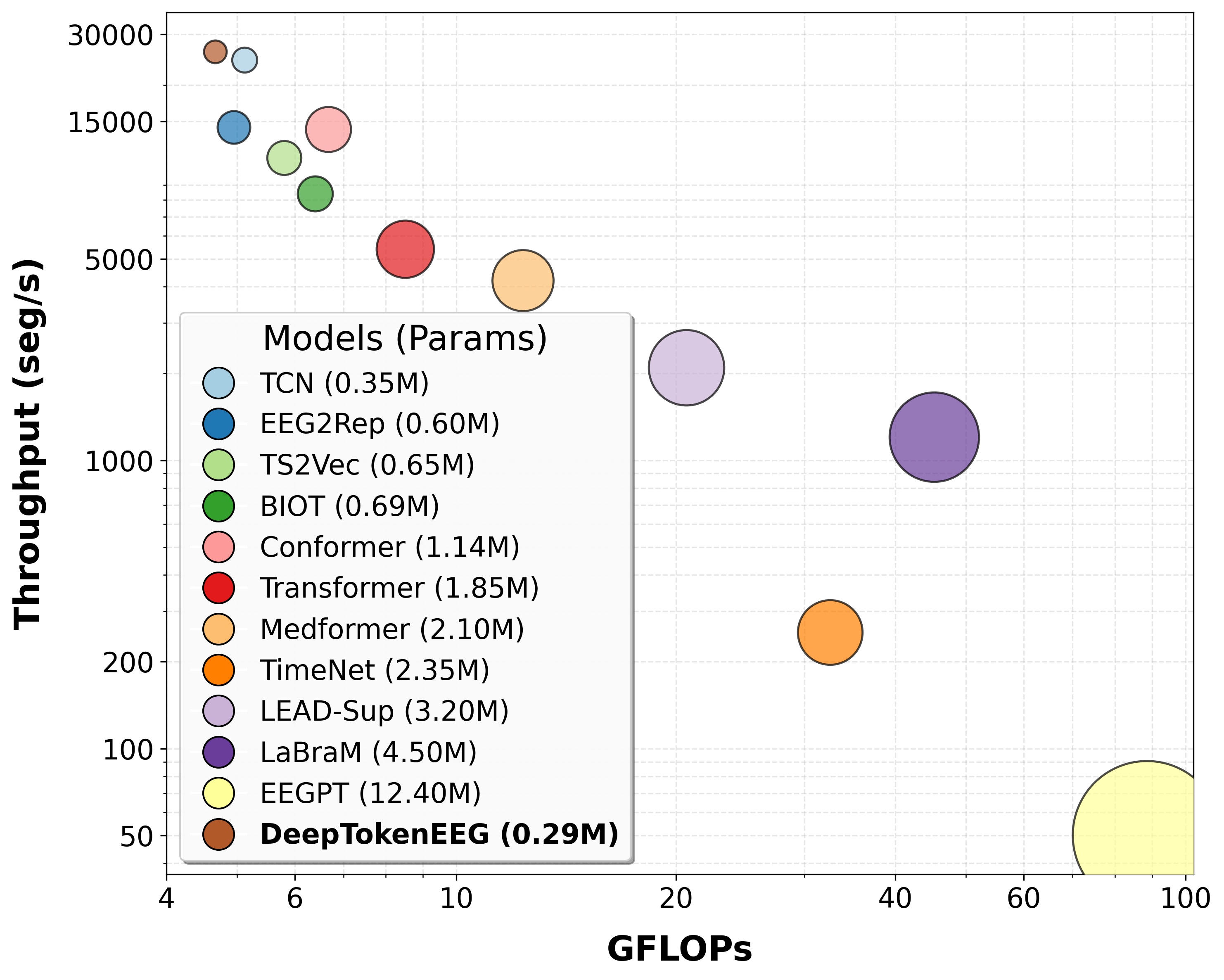}
\caption{\small\label{fig:performance}
Comparison of model complexity, throughput, and parameter size. Each marker represents a model, where the x-axis (log scale) shows GFLOPs and the y-axis (log scale) shows throughput. Marker size is proportional to the number of parameters, and colors/markers distinguish models. Models in the upper-left region indicate better efficiency. DeepTokenEEG achieves the best trade-off.}
\end{figure}

%

    

Finally, we evaluated the computational efficiency of the proposed method, a critical factor for real-world clinical deployment. Fig. \ref{fig:performance} presents a comparison of parameter counts, floating-point operations (GFLOPs), and inference throughput across various architectures. DeepTokenEEG achieves the most efficient performance across all metrics, requiring only 0.29 million parameters. This represents a substantial reduction compared to large-scale transformer baselines, such as LEAD (3.20M) and TimesNet (2.35M), being approximately 11.0 times and 8.1 times smaller, respectively.

Crucially, our method demonstrates superior computational speed and efficiency over even the most lightweight baselines. While previous lightweight models like EEG2Rep and TCN are optimized for efficiency, DeepTokenEEG still outperforms them, achieving the lowest computational cost (4.67 GFLOPs) and the highest throughput (26,173 segments/s). Specifically, it reduces the computational load compared to EEG2Rep (4.95 GFLOPs) and surpasses the inference speed of the fast TCN model (24,522 segments/s). These results indicate that the proposed depthwise tokenization strategy effectively eliminates redundancy while maximizing processing speed. The combination of high classification accuracy, minimal latency, and low resource consumption makes DeepTokenEEG highly suitable for integration into portable EEG headsets or edge computing devices.

\section{Discussion}
\label{Discussion}

Although various feature groups and classification models have been explored, most studies evaluate a limited combination of features and models, making it difficult to assess the true discriminative power of individual features. The lack of consistency in model-feature pairing further complicates comparisons. In addition, modern deep learning models require substantial computational resources, and early classification often relies on distinguishing AD from mild cognitive impairment (MCI) as a separate task. Therefore, this study aims to address these existing issues by achieving improved performance and providing a clearer classification of mild cognitive impairment.

The proposed DeepTokenEEG framework is predicated on the principle that robust Alzheimer’s diagnosis necessitates preserving the temporal stationarity of neural oscillations while simultaneously capturing multi-scale dependencies. Consequently, the selection of the four-level Stationary Wavelet Transform (SWT) is a critical design choice driven by the non-stationary nature of EEG signals. Unlike FFT, which sacrifices temporal localization, or Discrete Wavelet Transform (DWT), which suffers from shift-variance due to decimation, SWT provides a strictly translation-invariant representation. By eliminating downsampling, SWT preserves the precise temporal alignment of spectral features. This ensures that the downstream tokenization inputs remain stable against minor temporal shifts, effectively mitigating the aliasing artifacts that commonly degrade deep learning models trained on physiological signals.

The core novelty of DeepTokenEEG lies in its Embedding and Tokenization mechanism. Recognizing that neural degradation in AD manifests as subtle disruptions in both local high-frequency synchronization and long-range functional connectivity, we employ an architecture based on Tokenizer inspired by tokenizer in natural language processing, and a stack of Residual Blocks. To determine the optimal architecture, we also evaluate the number of Residual Block stages and conduct a comparative analysis between two configuration strategies: exponentially increasing dilation rates through each block ($d = 2^j$) versus constant dilation rates across stages. This design, inspired by Temporal Convolutional Networks (TCNs), offers distinct theoretical advantages over standard CNNs or RNNs: the exponential progression ($2^1, 2^2, 2^3$) allows the network’s effective receptive field to expand rapidly with depth. This is theoretically vital for capturing low-frequency oscillations (such as Delta and Theta rhythms)—key biomarkers for AD that require a wide temporal window to be fully characterized. This ensures the model learns long-term dependencies without the vanishing gradient problems associated with RNNs. In contrast to RNNs (LSTMs/GRUs) which are constrained by sequential processing, our dilated convolutional structure facilitates massively parallel processing on GPUs. This structural characteristic is the primary driver behind the model's superior inference speed.

Beyond theoretical rigor, DeepTokenEEG demonstrates exceptional computational efficiency. As evidenced by the comparative analysis, our model achieves State-of-the-Art (SOTA) performance with only 0.29 million parameters—approximately 11 times smaller than the LEAD framework and 8 times more efficient than TimesNet. This "lightweight yet powerful" characteristic addresses a critical gap in AI-assisted diagnosis: the need for models capable of operating on resource-constrained edge devices, such as portable EEG headsets, without reliance on heavy cloud computing.

Finally, the study shows that beyond alpha and beta waves, which are traditionally associated with the "slowing" of EEG rhythms in Alzheimer's pathology, the model demonstrates that gamma waves also serve as a significant feature for AD classification. This is evidenced by the high performance at the subject level, achieving an accuracy of $87.44\%$ and $91.55\%$ F1-Score. By effectively isolating and leveraging these specific frequency bands via SWT and Tokenization, DeepTokenEEG does not merely fit statistical patterns but captures biologically meaningful deviations in brain activity. This biological plausibility, combined with high generalization across datasets, suggests that DeepTokenEEG is a reliable candidate for scalable, non-invasive AD screening.

\section{Conclusion}
\label{Conclusion}

In this study, we presented DeepTokenEEG, a new deep learning model for Alzheimer’s disease detection using EEG signals. The main idea of our method is to use tokenization, through convolutional and depthwise tokenizers, to turn EEG signals into structured features. This design makes it possible to capture both the time patterns within each channel and the relationships across channels, while keeping the model small and efficient.

Through experiments on multiple datasets, DeepTokenEEG showed strong results compared to existing approaches. It achieved better and comparable performance to state-of-the-art methods such as LEAD, while using far fewer parameters. The model was effective at both the sample level and the subject level, showing that it can provide stable predictions even when the data comes from different sources.

Overall, DeepToken offers a practical and lightweight solution for EEG-based Alzheimer’s diagnosis. This token-based representation helps lower feature-level noise, improves robustness through data augmentation, and allows better adaptation across datasets. However,the evaluation has so far been limited to a few EEG datasets, and broader testing on larger and more diverse datasets is needed to confirm its generalizability. In addition, the framework has mainly been applied to Alzheimer’s detection, and its potential for other neurological disorders has yet to be investigated.

Future work will aim to extend our work by conducting experiments under two specific directions. The first will focus on differentiating Alzheimer’s disease (AD) from non-AD conditions within a multi-class classification framework, leveraging datasets including AD, healthy controls (HC), and other neurological disorders such as frontotemporal dementia (FTD), Parkinson’s disease (PD), multiple sclerosis (MS), and epilepsy. This direction will enable more comprehensive classification by explicitly capturing the boundary between AD and non-AD subjects. The second will manage a classification problem involving AD, HC, and other brain disorders not categorized as AD (OD-AD). This will facilitate more accurate differentiation between these groups, thereby contributing to a deeper understanding of AD-related mechanisms and supporting earlier and more precise intervention strategies.  

\bibliographystyle{ieeetr}
\bibliography{IEEEabrv, References}

\end{document}